\documentclass[10pt,twocolumn,letterpaper]{article}

\usepackage[pagenumbers]{cvpr} 

\usepackage{graphicx}
\usepackage{amsmath}
\usepackage{amssymb}
\usepackage{booktabs}
\usepackage{multirow}

\usepackage[pagebackref,breaklinks,colorlinks]{hyperref}

\usepackage{color}

\usepackage[capitalize]{cleveref}
\crefname{section}{Sec.}{Secs.}
\Crefname{section}{Section}{Sections}
\Crefname{table}{Table}{Tables}
\crefname{table}{Tab.}{Tabs.}

\def\cvprPaperID{8954} 
\def\confName{CVPR}
\def\confYear{2022}

\begin{document}

\title{Label Relation Graphs Enhanced Hierarchical Residual Network for Hierarchical Multi-Granularity Classification}

\author{Jingzhou Chen\\
Zhejiang University\\
\and
Peng Wang\\
Zhejiang University\\
\and
Jian Liu\\
Ant Group\\
\and
Yuntao Qian\thanks{Corresponding author}\\
Zhejiang University
}
\maketitle

\begin{abstract}
Hierarchical multi-granularity classification (HMC) assigns hierarchical multi-granularity labels to each object and focuses on encoding the label hierarchy, \eg, [``Albatross", ``Laysan Albatross"] from coarse-to-fine levels. However, the definition of what is fine-grained is subjective, and the image quality may affect the identification. Thus, samples could be observed at any level of the hierarchy, \eg, [``Albatross"] or [``Albatross", ``Laysan Albatross"], and 
examples discerned at coarse categories are often neglected in the conventional setting of HMC. In this paper, we study the HMC problem in which objects are labeled at any level of the hierarchy. The essential designs of the proposed method are derived from two motivations: (1) learning with objects labeled at various levels should transfer hierarchical knowledge between levels; (2) lower-level classes should inherit attributes related to upper-level superclasses.  
The proposed combinatorial loss maximizes the marginal probability of the observed ground truth label by aggregating information from related labels defined in the tree hierarchy. If the observed label is at the leaf level, the combinatorial loss further imposes the multi-class cross-entropy loss to increase the weight of fine-grained classification loss. Considering the hierarchical feature interaction, we propose a hierarchical residual network (HRN), in which granularity-specific features from parent levels acting as residual connections are added to features of children levels. Experiments on three commonly used datasets demonstrate the effectiveness of our approach compared to the state-of-the-art HMC approaches and fine-grained visual classification (FGVC) methods exploiting the label hierarchy. 
\end{abstract}

\section{Introduction}
\label{sec:intro}

\begin{figure}
  \centering
  \begin{subfigure}{0.98\linewidth}
    \centering
    \includegraphics[width=1.0\linewidth]{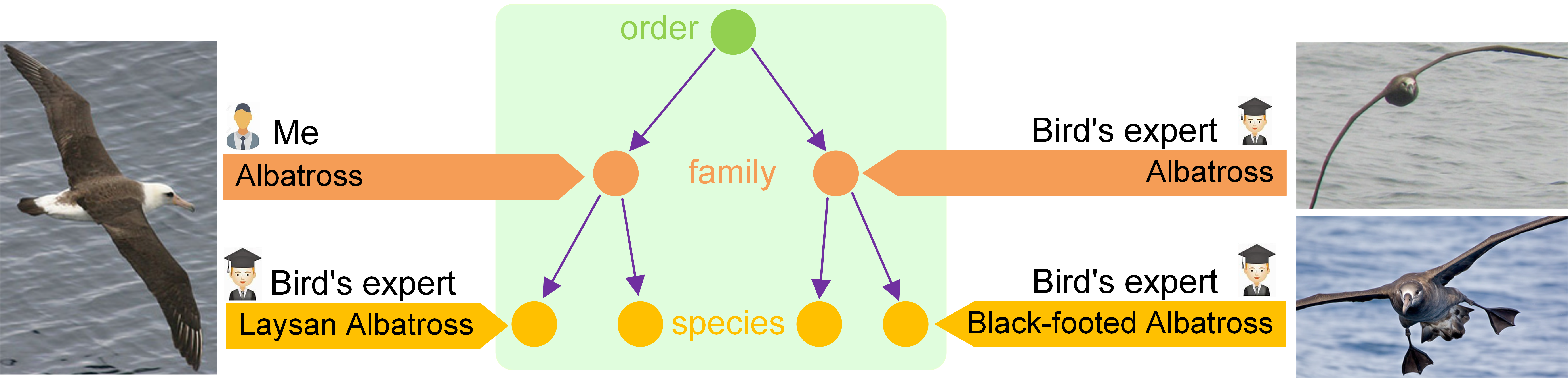}
    \caption{Differences in domain knowledge and interference from the image occlusion.}
    \label{fig:Moti1}
  \end{subfigure}

  \begin{subfigure}{0.98\linewidth}
    \centering
    \includegraphics[width=1.0\linewidth]{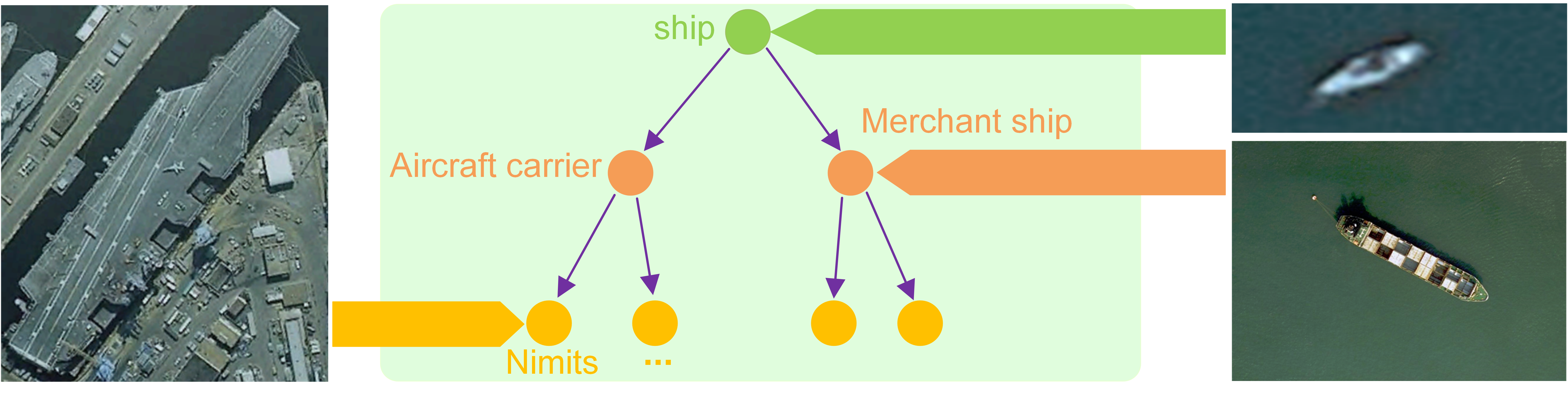}
    \caption{Large variations of image resolutions.}
    \label{fig:Moti2}
  \end{subfigure}
  \caption{Different objects can be discerned at various levels in the label hierarchy due to differences in domain knowledge or image quality such as occlusion or resolution.}
  \label{fig:Moti}
\end{figure}

Traditional single-granularity classification usually assigns a single label to a given object from a set of mutually exclusive class labels. For instance, FGVC aims at distinguishing objects from different subordinate-level categories within a given object category, e.g., subcategories of birds~\cite{wah2011caltech}, cars~\cite{krause20133d}, aircraft~\cite{maji2013fine}.  
However, the definition of what is fine-grained is subjective, and the image quality may affect the identification, as illustrated in~\cref{fig:Moti}. A bird can be discerned as Albatross or Laysan Albatross due to differences in domain knowledge. Moreover, a bird expert recognizes a bird as Albatross rather than Black-footed Albatross because of the occlusion of key parts. Airborne or satellite image resolutions often have large variations, causing objects to be recognized at different levels. 
These challenges increase the difficulty of constructing a dataset for single-granularity classification, while images annotated as coarse categories are also overlooked.

Compared to single-granularity classification, a more preferable solution is to employ hierarchical multi-granularity labels to describe an object, which provides more flexible options for annotators with different knowledge backgrounds~\cite{chang2021your}. HMC~\cite{silla2011survey} aims to exploit hierarchical multi-granularity labels and embeds the label hierarchy in loss function or network architecture. Whereas conventional HMC usually evaluates each sample with complete hierarchical labels from the coarsest to the finest granularity. A more robust HMC model should effectively utilize examples observed at various levels in the hierarchy, \eg, making use of bird images annotated as [``Albatross"] and [``Albatross", ``Laysan Albatross"].

In this paper, we study the HMC problem in which samples are labeled at any level of the hierarchy.
We factorize this problem into two aspects: (1) how to effectively use instances labeled at different levels; (2) how to perform hierarchical feature interaction in the network architecture. 
For the first problem, we adopt a tree hierarchy that defines two kinds of semantic relationships between labels: parent-child correlations between levels and mutual exclusion at the same level. 
Inspired by the work of~\cite{deng2014large}, if an instance is discerned at a label in the hierarchy, we maximize its marginal probability in the probability space constrained by the tree hierarchy. Such marginalization enjoys two benefits: learning with the coarse-level label could impact decisions of fine-grained subclasses while learning with the fine-level label aids the prediction of coarse-grained superclasses.
Moreover, if the ground truth label is observed at the leaf level, we further impose the multi-class cross-entropy loss to enhance the discriminative power among fine-grained categories.

\begin{figure*}
  \centering
  \includegraphics[width=0.88\linewidth]{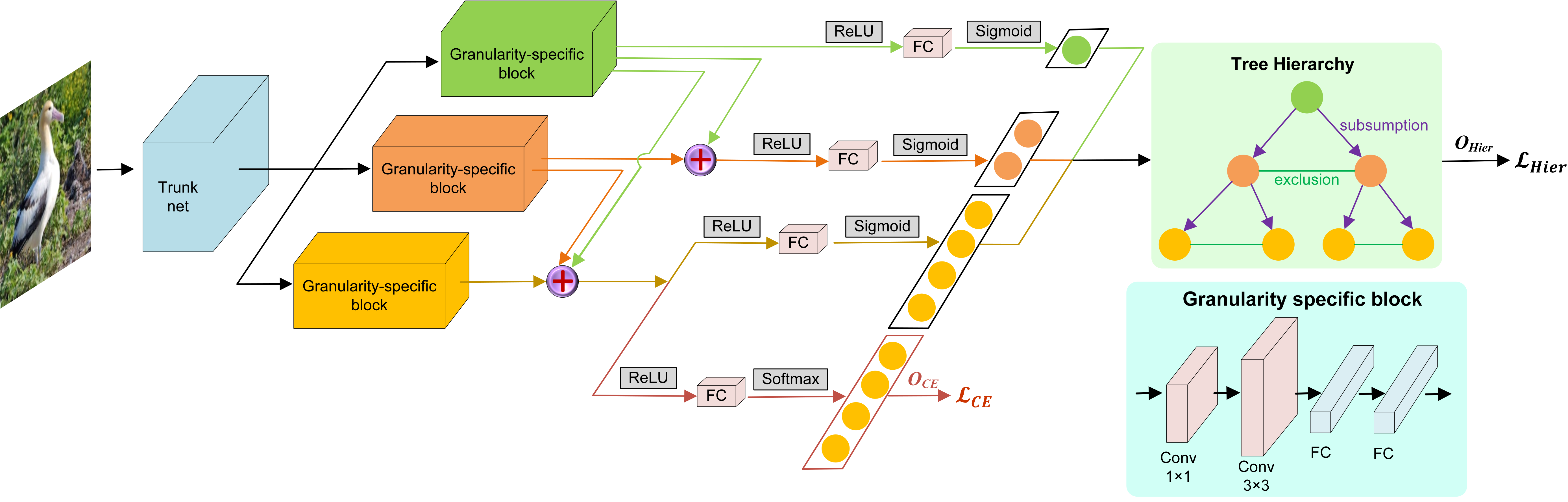}
  \caption{The network architecture consists of a trunk network (ResNet-50), hierarchical feature interaction module, and two parallel output channels: $O_{Hier}$ and $O_{CE}$ forming the probabilistic classification loss ($\mathcal{L}_{Hier}$) and the cross-entropy loss ($\mathcal{L}_{CE}$), respectively. We illustrate the network architecture on CUB-200-2011 dataset that contains three hierarchical levels. Granularity-specific block for each hierarchical level process feature maps generated from the trunk network, then these hierarchical features interact via residual connections, \ie, features from parent levels acting as skip connections are added to features of children levels. $O_{Hier}$ organizes sigmoid outputs from three hierarchical levels using the tree hierarchy, and $O_{CE}$ generates softmax outputs corresponding to the fine-grained leaf categories.}
  \label{fig:overview}
\end{figure*}

Another critical issue is to design appropriate hierarchical feature interaction that reflects the label hierarchy. A distinct characteristic of hierarchical categories is that from coarse-to-fine levels, fine-level classes not only have unique attributes but also inherit attributes related to coarse-level superclasses. Based on this property, we propose a hierarchical residual network (HRN) illustrated in~\cref{fig:overview}. We first set up granularity-specific layers to disentangle hierarchical features from the trunk network. Then, these hierarchical features interact via residual connections~\cite{he2016identity,xie2017aggregated,huang2017densely,howard2017mobilenets,howard2019searching,ma2018shufflenet,vaswani2017attention}, \ie, features from parent levels acting as skip connections are added to features of children levels.
Experiments on three commonly used FGVC datasets demonstrate the effectiveness of our approach compared to the state-of-the-art HMC approaches and FGVC methods exploiting hierarchical knowledge under two evaluation metrics~\cite{vens2008decision}. 

\section{Related Work}
\label{sec:related}

\subsection{Hierarchical Multi-Granularity Classification}

HMC problems naturally arise in many domains, such as text categorization~\cite{lewis2004rcv1,rousu2006kernel,mayne2009hierarchically} and functional genomics~\cite{barutcuoglu2006hierarchical,vens2008decision,schietgat2010predicting}. 
In text categorization, an increasing number of works~\cite{mao2019hierarchical,huang2019hierarchical,meng2019weakly,chen2020hyperbolic} leveraged the label hierarchy to improve accuracy. 
In image classification, HMC systems have been used to annotate medical images~\cite{dimitrovski2011hierarchical} and classify diatom images~\cite{dimitrovski2012hierarchical}. 
Based on deep neural networks (DNNs), the studies usually go along two paths: 
mapping the label hierarchy to network architectures~\cite{cerri2016reduction,cerri2014hierarchical,peng2018large,wehrmann2018hierarchical} or loss functions that impose the hierarchical constraints~\cite{deng2014large,giunchiglia2020coherent}. 
HMC with local multi-layer perceptrons (HMC-LMLP)~\cite{cerri2016reduction} proposed to train a chain of multi-layer perceptron (MLP) networks, each corresponding to a hierarchical level. The input of each MLP uses the output provided by the previously trained MLP to augment the feature vector of the instance. This supervised incremental greedy procedure continues until the last level of the hierarchy is reached. HMC network (HMCN)~\cite{wehrmann2018hierarchical} comprised multiple local outputs, with one local output layer per hierarchical level of the class hierarchy plus a global output layer that captures the cumulative relationships forwarded across the entire network. All local outputs are then concatenated and pooled with the global output to generate a final consensual prediction. 
HMC-LMLP and HMCN embed label hierarchy in their network architecture. Their loss functions sum over binary cross-entropy losses from each hierarchical level, which assumes each label is independent of each other, causing the implicit hierarchical relations between two semantic labels to be ignored.

Another line of HMC works encodes the label hierarchy in loss functions by imposing the hierarchical constraints. 
Coherent HMC neural network (C-HMCNN)~\cite{giunchiglia2020coherent} revised the binary cross-entropy loss to satisfy the parent-child constraint. The revision ensures that no hierarchy violation happens, \ie, for any threshold, when C-HMCNN predicts a sample belonging to a class, this sample also belongs to its parent classes. Moreover, C-HMCNN can teach the network how to better make the prediction on the higher level classes using the prediction results on the lower level ones. 
While C-HMCNN only restricts the parent-child correlation, other kinds of semantic relations between hierarchical labels can be constructed using graphs. 
Deng \etal~\cite{deng2014large} formalized semantic connections between any two labels into a directed acyclic graph (DAG). They built a modified junction tree algorithm that contains multiple loops during message passing on the junction tree to compute the probabilistic classification loss defined on the DAG. 

\subsection{Fine-Grained Visual Classification}

Since FGVC inherently forms a hierarchy with different levels of concept abstraction, many approaches~\cite{zhang2016embedding,zhou2016fine,shi2018fine,chen2018fine,chang2021your} proposed to exploit the hierarchical label structure of FGVC. 
Zhang \etal~\cite{zhang2016embedding} generalized the triplet loss by describing inequalities of the distance between images belonging to the same fine-grained class, different fine-grained classes but the same coarse class, and different coarse classes. Shi \etal~\cite{shi2018fine} proposed a generalized large-margin loss that not only reduces between-class similarity and within-class variance of the learned features but also makes the subclasses belonging to the same coarse class be more similar than those belonging to different coarse classes in the feature space. Chen \etal~\cite{chen2018fine} developed a novel hierarchical semantic embedding framework that incorporates the predicted score vector of the higher level as prior knowledge to learn finer-grained feature representation at each hierarchical level. During training, the predicted score vector of the higher level is also employed to regularize sub-categories prediction by using it as soft targets. Chang \etal~\cite{chang2021your} leveraged level-specific classification heads to disentangle coarse-level features with fine-grained ones and allowed fine-grained features to participate in coarser-grained label predictions but constraining the gradient flow to only update the parameters within each classification head. Their method reaches the state-of-the-art results on the traditional single-label FGVC problem. 
These approaches refine the feature representation related to hierarchical levels in the feature space. 
They developed the loss function based on the multi-class cross-entropy loss that implies the mutual exclusion among classes at the same hierarchical level. 
However, they neglect to encode other label relations like the parent-child correlation to transfer hierarchical knowledge between levels using samples observed at different levels. 


\section{Proposed Methods}
\label{sec:methods}

\subsection{Network Architecture}
\label{sec:network}
 
Our network architecture includes a trunk network, a hierarchical feature interaction module, and two parallel output channels, see Fig. \ref{fig:overview}. 
The trunk network is used to extract features from the input images and any common network is applicable. Here, we adopt the ResNet-50 as the trunk network since it is widely used for feature extraction. 
The hierarchical feature interaction module contains granularity-specific block and residual connections. These blocks share the same structure that comprises two convolutional layers and two fully connected (FC) layers. Each block is designed to extract the specialized feature for one hierarchical level. 
The residual connections first linearly combine features of fine-level subclasses with features of coarse-level superclasses. Accordingly, subclasses not only have unique attributes but also inherit the attributes from their superclasses. Then, non-linear transformation (ReLU) is applied to combined features. 

We set up two output channels in our model. The first output channel is utilized to compute the probabilistic classification loss based on the tree hierarchy, in which each sigmoid node corresponds to a distinct label in the hierarchy. We perform the non-linear projection by sigmoid instead of softmax because sigmoid reflects the independent relations, whereas softmax implies mutual exclusion. The sigmoid nodes from each hierarchical level are then organized with the tree hierarchy to comply with the hierarchical constraints. 
The second output channel computes the multi-class cross-entropy loss imposed on the leaf level so that the mutually exclusive fine-grained classes gain more attention during training. For simplicity, we denote the first and the second output channel as $O_{Hier}$ and $O_{CE}$, respectively. 

\subsection{Loss Function}
\label{sec:loss}

The proposed combinatorial loss integrates two forms of losses: the probabilistic classification loss and the multi-class cross-entropy loss. We first formalize the tree hierarchy to encode semantic relations between hierarchical labels. The probabilistic classification loss defined on the tree hierarchy aims to transfer hierarchical knowledge during training. We empirically find that if the training samples labeled at the leaf level are few, the probabilistic classification loss fails to well separate the fine-grained leaf classes. One simple but feasible solution is to increase the weight of fine-grained classification loss. Therefore, we further impose the multi-class cross-entropy loss on the leaf categories, which obeys the mutually exclusive constraint among fine-grained classes defined in the tree hierarchy.

\subsubsection{The Formalism of Tree Hierarchy}
\label{sec:tree}

The tree hierarchy $G = (V, E_h, E_e)$ consists of a set of nodes $V = \{v_1, \ldots, v_n\}$, directed edges $E_h \subseteq V \times V$, and undirected edges $E_e \subseteq V \times V$. Each node $v \in V$ corresponds to a distinct class label. The number of nodes $n$ equals the number of all labels in the hierarchy. A directed edge $(v_i, v_j) \in E_h$ is a subsumption edge, indicating that class $i$ subsumes label $j$, \eg, Albatross is a parent or superclass of Black-footed Albatross. An undirected edge $(v_i, v_j) \in E_e$ is an exclusion edge, denoting that classes $v_i$ and $v_j$ are mutually exclusive, \eg, a bird cannot be the Black-footed Albatross and Laysan Albatross simultaneously. Any two nodes share a subsumption edge or an exclusion edge.

Each class label takes binary values, \ie, $v_i \in \{0,1\}$, representing whether an object belongs to this class or not. Each edge then defines a constraint on the binary values that two labels of its incident nodes can take. An assignment of $(v_i, v_j) = (0, 1)$ (e.g. a Black-footed Albatross but not a Albatross) for a subsumption edge $(v_i, v_j) \in E_h$ is illegal, while $(v_i, v_j) = (1, 1)$ (it is both Black-footed Albatross and Laysan Albatross) is also an illegal assignment for an exclusion edge $(v_i, v_j) \in E_e$. Defined by these local constraints of individual edges, a legal global assignment of all labels in the hierarchy is a binary label vector $\mathbf{y} \in \{0, 1\}^n$ for an object. The set of all legal global assignments forms the state space $S_G \subseteq \{0, 1\}^n$ of tree $G$. We can infer $S_G$ to be a matrix $\mathbf{S} \in \mathcal{R}^{(n+1) \times n}$, where each row represents a legal binary label vector $\mathbf{y}$. We traverse all legal assignments by assigning each label a value of $1$, along with an assignment that is all zeros.

\subsubsection{Probabilistic Classification Loss}\label{ssec:hex_loss}

We calculate the probabilistic classification loss from $O_{Hier}$, and each sigmoid node in $O_{Hier}$ corresponds to a class label in the tree hierarchy. Suppose the number of sigmoid nodes is $n$, and $\mathbf{y} \in \{0, 1\}^n$ is the binary label vector representing an assignment of all labels. Given an input image $\mathbf{x}$, the joint probability of all sigmoid nodes concerning the assignment $\mathbf{y}$ can be computed as:
\begin{equation}
\tilde{P}(\mathbf{y}|\mathbf{x}) = \prod^n_{i=1}\phi_i(\bar{x}_i, y_i)\prod_{i,j\in\{1,\ldots,n\}}\psi_{i,j}(y_i, y_j)
\end{equation}
where $\bar{x}_i$ is the sigmoid output of the $i$-th label node, $\tilde{P}(\mathbf{y}|\mathbf{x})$ is the unnormalized probability, and $\phi_i(\bar{x}_i, y_i) = e^{\bar{x}_i[y_i=1]}$. $\psi_{i,j}(y_i, y_j)$ is the constraint defined in the tree hierarchy between any two labels in $\mathbf{y}$:
\begin{equation}
\psi_{i,j}(y_i, y_j) =
  \begin{cases}
    0, & \text{if violates constraints} \\
    1, & \text{otherwise}
  \end{cases}
\end{equation}
The joint probability is then normalized by $Pr(\mathbf{y}|\mathbf{x}) = \frac{\tilde{P}(\mathbf{y}|\mathbf{x})}{Z(\mathbf{x})}$, where $Z(\mathbf{x})$ is the partition function that sums over all legal assignments $\bar{\mathbf{y}} \in S_G$ in the state space of tree $G$:
\begin{equation}
Z(\mathbf{x}) = \sum_{\bar{\mathbf{y}}\in\{0, 1\}^n}\prod^n_{i=1}\phi_i(\bar{x}_i, \bar{y}_i)\prod_{i,j\in\{1,\ldots,n\}}\psi_{i,j}(\bar{y}_i, \bar{y}_j)
\end{equation}

If input image $\mathbf{x}$ is observed at the $i$-th label in the tree hierarchy, \ie, $y_i=1$, we can obtain the marginal probability $Pr(y_i=1|\mathbf{x})$ of label $i$ by summing over all legal assignments $\bar{\mathbf{y}} \in S_G$ that include $\bar{y}_i=1$:
\begin{equation}
Pr(y_i=1|\mathbf{x})=\frac{1}{Z(\mathbf{x})}\sum_{\bar{\mathbf{y}}:\bar{y}_i=1}\prod_i\phi_i(\bar{x}_i, \bar{y}_i)\prod_{i,j}\psi_{i,j}(\bar{y}_i, \bar{y}_j)
\end{equation}
The marginal probability of a leaf label in tree $G$ relies on the sum of its ancestors' scores because all its ancestors must be 1 if the label of this leaf node takes value 1, which enables the parents' scores to impact the descendants' decisions. On the other hand, the marginal probability of a parent label is marginalized over all possible states of its descendants, \ie, aggregating the information from all its subclasses.

We propose to compute marginalization via matrix multiplication. Suppose the network outputs $\mathbf{X} \in \mathcal{R}^{n \times k}$ from $O_{Hier}$, where $n$ is the number of sigmoid nodes, and $k$ stands for batch size. Each column in $\mathbf{X}$ is the output vector corresponding to a sample in the batch. The unnormalized joint probability can be computed as $\mathbf{J} = \exp(\mathbf{S}\mathbf{X})$, and the partition function $\mathbf{z}$ can be calculated by summing each column of $\mathbf{J}$. To obtain the marginal probability of the $j$-th sample labeled at $i$, we first search for eligible rows in the $i$-th column of $\mathbf{S}$ that qualify $\mathbf{S}[:, i] > 0$, then we sum the corresponding elements in the $j$-th column of $\mathbf{J}$, finally, we normalize the summation by dividing the $j$-th element in $\mathbf{z}$.

In the training process, the observed label can be at any level of the hierarchy, and we maximize the marginal likelihood of the observed ground truth label.
Given $m$ training samples $\mathcal{D}=\{\mathbf{x}^{(l)}, \mathbf{y}^{(l)}, g^{(l)}\}$, $l=1,\ldots,m$, where $\mathbf{y}^{(l)}$ is the complete ground truth label vector and $g^{(l)} \in \{1,\ldots,n\}$ is the index of the observed label, the probabilistic classification loss is defined as:
\begin{equation}
\mathcal{L}_{Hier}(\mathcal{D}) = - \frac{1}{m}\sum_l^m \ln(Pr(y^{(l)}_{g^{(l)}}=1|\mathbf{x}^{(l)}))
\end{equation}

\subsubsection{Combinatorial Loss}\label{ssec:com_loss}

The multi-class cross-entropy loss is commonly used in FGVC to separate fine-grained categories. We add $\mathcal{L}_{CE}$ to our model to further increase the discriminative power for fine-grained leaf classes. $\mathcal{L}_{CE}$ employs softmax outputs from $O_{CE}$, in which each node corresponds to a fine-grained leaf label in the tree hierarchy. Softmax outputs imply mutually exclusive relations among fine-grained classes, which is consistent with the hierarchy constraint defined in the tree hierarchy. We combine $\mathcal{L}_{CE}$ with $\mathcal{L}_{Hier}$ as follows:
\begin{equation}
\mathcal{L}_{com}(\mathbf{x}^{(l)},y^{(l)}_{g^{(l)}}) =
  \begin{cases}
    \mathcal{L}_{CE} + \mathcal{L}_{Hier}, &\text{if } g^{(l)} \text{ is in}\\
     &\text{leaf nodes} \\
    \mathcal{L}_{Hier}, &\text{otherwise}
  \end{cases}
\end{equation}
Depending on whether $\mathbf{x}^{(l)}$ is labeled at fine-grained leaf categories, the combined loss decides whether it needs to incorporate $\mathcal{L}_{CE}$ or not. Finally, the total loss on $\mathcal{D}$ is:
\begin{equation}
\mathcal{L}_{total}(\mathcal{D}) = \sum_l \mathcal{L}_{com}(\mathbf{x}^{(l)},y^{(l)}_{g^{(l)}})
\end{equation}

\section{Experiments}
\label{sec:exper}

\subsection{Implementaion Details}\label{ssec:implement}

In all our experiments, we resize input images to $448 \times 448$ and train every single experiment for 200 epochs. Random horizontal flipping and random cropping (random cropping for training and center cropping for testing) are applied for data augmentation. We adopt ResNet-50 pre-trained on ImageNet as our trunk network and use stochastic gradient descent (SGD) with a momentum of 0.9, weight decay of 0.0005 to optimize our model. The batch size is set to 8. Meanwhile, the learning rates of the convolution layers and the FC layers newly added for hierarchical interaction are initialized as 0.002 and adjusted by the cosine annealing strategy~\cite{loshchilov2016sgdr}. The learning rates of the trunk layers are maintained as 1/10 of the newly added layers. The code will be made publicly accessible. 

\subsection{Datasets and Experimental Designs}\label{ssec:data}

We evaluate our proposed method on three widely used FGVC datasets, \ie, CUB-200-2011~\cite{wah2011caltech}, Aircraft~\cite{maji2013fine}, and Stanford Cars~\cite{krause20133d}. However, CUB-200-2011 and Stanford Cars only provide one fine-grained label for each image. To construct a taxonomy of label hierarchy for these two datasets, we learn from the work of Chang \etal~\cite{chang2021your}, in which they trace parent nodes in Wikipedia pages. CUB-200-2011 is the most widely used benchmark for FGVC. It covers 11,788 bird images that are re-organized into a three-level label hierarchy with 13 orders, 38 families, and 200 species. Aircraft contains 10,000 images consisting of a three-level label hierarchy with 30 makers, 70 families, and 100 models. Stanford Cars contains 16,185 images of 196 classes of cars. It is re-organized into a two-level label hierarchy with 9 car types and 196 specific models. We do not use any bounding box/part annotations in all our experiments and adopt the official train and test splits for evaluation. 

Besides assigning hierarchical multi-granularity labels for each image, experimental designs simulate the aforementioned situation where samples are observed at different levels of the hierarchy.
To imitate the lack of domain knowledge, we select 0\%, 30\%, 50\%, 70\%, and 90\% samples from each fine-grained class and relabel their last-level fine-grained classes to immediate parent classes in the training set, respectively. Considering the impact of image quality, we conduct another experiment by reducing the image resolution of selected samples using the nearest-neighbor interpolation with a factor of 4 after relabeling. The extreme case 0\% represents the conventional setting of HMC or fine-grained classification that exploits the label hierarchy. Other cases indicate that part of samples are observed at internal levels of the tree hierarchy, and the rest owns the complete label hierarchy from the highest level to the lowest fine-grained level. All images in the test set are tested with the complete label hierarchy. 

\subsection{Evaluation Metrics}\label{ssec:eval}

To reasonably evaluate the performance of HMC on FGVC datasets, we employ two evaluation metrics. The first metric follows the convention of FGVC and uses the overall accuracy (OA). The output of HMC models is a probability vector for each class. Concerning the hierarchical label structure, we take the maximum value of the output probability vector corresponding to each hierarchical level as the predicted label and compute OA on the test set. The second criterion commonly used in HMC literature~\cite{vens2008decision,wehrmann2018hierarchical,giunchiglia2020coherent} measures the area under the average precision and recall curve $AU(\overline{PRC})$. Instead of calculating the precision and recall curve (PRC) for each class, $AU(\overline{PRC})$ computes an average PRC to evaluate the output probability vector of all classes in the hierarchy. Specifically, for a given threshold value, one point $(\overline{Prec}, \overline{Rec})$ in the average PRC is computed as:
\begin{equation}
\begin{aligned}
\overline{Prec} & = \frac{\sum^n_{i=1}TP_i}{\sum^n_{i=1}TP_i+\sum^n_{i=1}FP_i}\\
\overline{Rec} & = \frac{\sum^n_{i=1}TP_i}{\sum^n_{i=1}TP_i+\sum^n_{i=1}FN_i}
\end{aligned}
\end{equation}
where $i$ ranges over all classes, and $TP_i$, $FP_i$, and $FN_i$ are the numbers of true positives, false positives, and false negatives for class label $i$, respectively. By varying the threshold, an average PRC is obtained and $AU(\overline{PRC})$ denotes the area under this curve. $AU(\overline{PRC})$ also has the advantage of being independent of the threshold used to predict when a sample belongs to a particular class (which is often heavily application-dependent).

\subsection{Ablation Study}\label{ssec:abla} 

In this section, we first conduct ablation studies to investigate two key designs of the proposed method on CUB-200-2011: HRN and combinatorial loss. 
Then, we conduct visualization experiments by illustrating the attention regions for each hierarchical level using Grad-Cam~\cite{selvaraju2017grad} to demonstrate the role of HRN and combinatorial loss. 

\subsubsection{Significance of HRN}\label{ssec:hrn}

As displayed in~\cref{fig:overview}, we analyze three components of HRN: granularity-specific block (GSB), the linear combination of hierarchical features (LC), and non-linear transformation of combined features (ReLU). We report OA on the species level with the relabeling proportion of 0\% in~\cref{tab:hrn}. The model that only contains ResNet-50 and combinatorial loss obtained a result of 84.32. As more components of HRN are integrated into the model, we gradually achieve better results.

\begin{table}
  \centering\footnotesize
  \caption{OA on the species level with the relabeling proportion of 0\% by gradually adding each component in HRN: granularity-specific block (GSB), the linear combination of hierarchical features (LC), and non-linear transformation of combined features (ReLU).}
  \label{tab:hrn}
  \begin{tabular}{lc}
    \toprule
    Component & OA \\
    \midrule
    Combinatorial Loss & 84.32 \\
    Combinatorial Loss + GSB & 85.77 \\
    Combinatorial Loss + GSB + LC & 86.17 \\
    Combinatorial Loss + GSB + LC + ReLU & \textbf{86.60} \\
    \bottomrule
  \end{tabular}
\end{table}

\subsubsection{Contribution of Combinatorial Loss}\label{ssec:loss}

In this subsection, we validate the effectiveness of combining the probabilistic classification loss ($\mathcal{L}_{Hier}$) with the multi-class cross-entropy loss ($\mathcal{L}_{CE}$). \cref{tab:loss} records OA on the species level with five relabeling proportions. In~\cref{tab:loss}, it can be found that when more training samples are relabeled to coarse-grained classes, the fine-grained classification performance of $\mathcal{L}_{Hier}$ degenerates drastically. In contrast, the combinatorial loss consistently outperforms $\mathcal{L}_{Hier}$ by adding $\mathcal{L}_{CE}$ imposed on fine-grained leaf classes.

\begin{table}
  \centering\footnotesize
  \caption{OA on the species level by analyzing the effectiveness of combining the probabilistic classification loss ($\mathcal{L}_{Hier}$) with the multi-class cross-entropy loss ($\mathcal{L}_{CE}$).}
  \label{tab:loss}
  \begin{tabular}{cccccc}
    \toprule
    Relabeling & 0\% & 30\% & 50\% & 70\% & 90\% \\
    \midrule
    $\mathcal{L}_{Hier}$ & 84.56 & 76.66 & 64.36 & 45.10 & 28.69 \\
    $\mathcal{L}_{Hier}$ + $\mathcal{L}_{CE}$ & \textbf{86.60} & \textbf{83.91} & \textbf{80.52} & \textbf{73.96} & \textbf{53.02} \\
    \bottomrule
  \end{tabular}
\end{table}

\subsubsection{Visualization Experiments}\label{ssec:cam}

\begin{figure}
  \centering
  \begin{subfigure}{0.98\linewidth}
    \centering
    \includegraphics[width=1.0\linewidth]{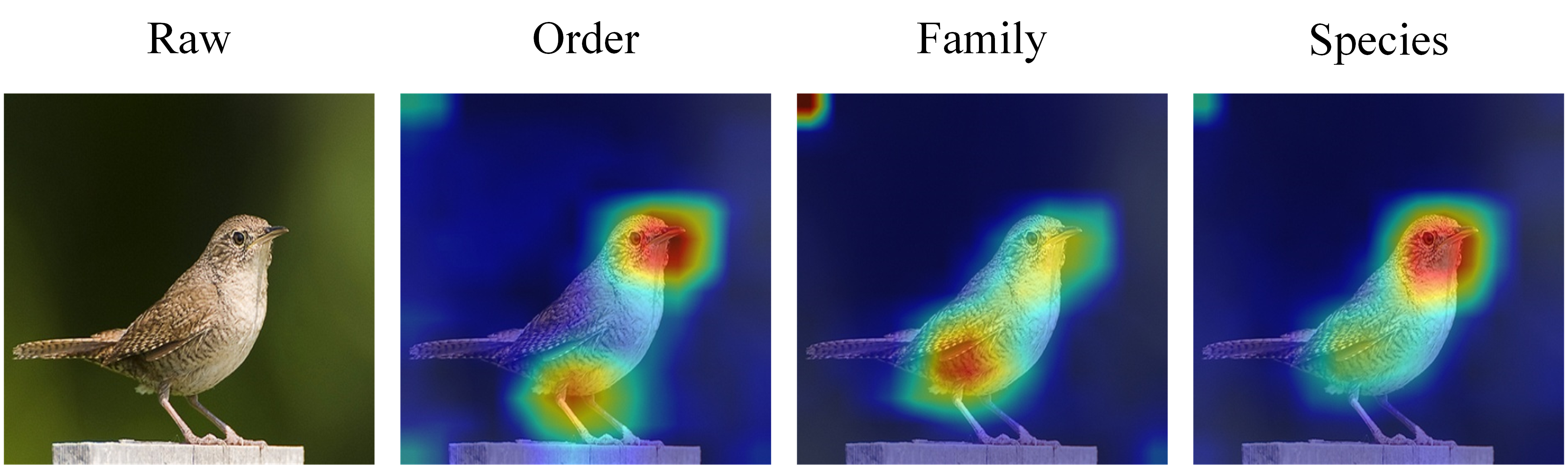}
    \caption{House Wren}
    \label{fig:196}
  \end{subfigure}

  \begin{subfigure}{0.98\linewidth}
    \centering
    \includegraphics[width=1.0\linewidth]{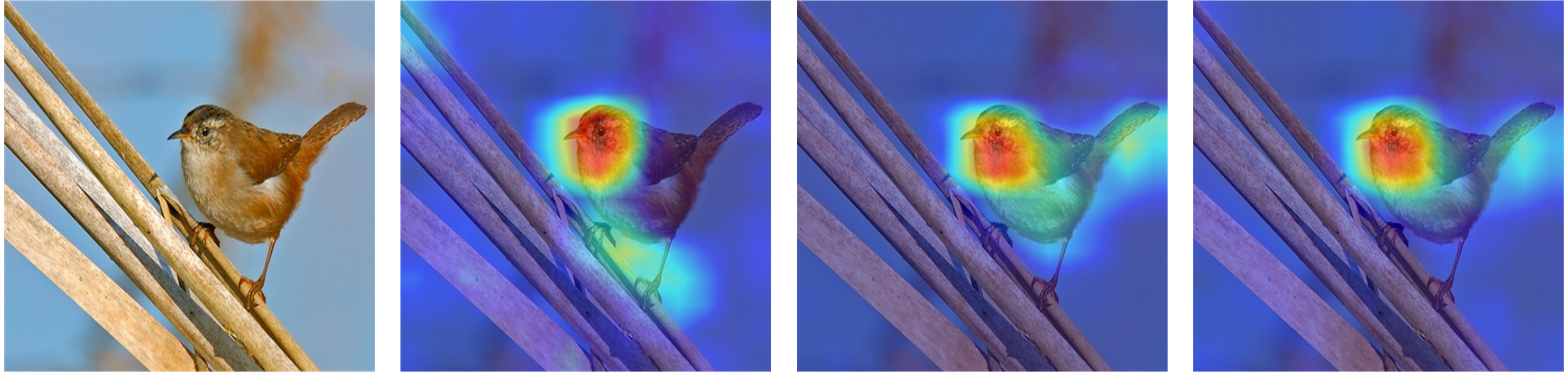}
    \caption{Marsh Wren}
    \label{fig:197}
  \end{subfigure}
  \caption{Visual attention maps of two species of birds from the same family (Troglodytidae) and order (Passeriformes).}
  \label{fig:Cam}
\end{figure}

We conduct visualization experiments to demonstrate that granularity-specific blocks can capture different regions of interest while hierarchical knowledge can be transferred across levels via hierarchical residual interaction and supervision of combinatorial loss. To this end, we adopt Grad-Cam to visualize different attention regions of each hierarchical level by propagating their respective gradients back to feature maps generated from the trunk network. 
\cref{fig:Cam} illustrates two species of birds: House Wren and Marsh Wren from the same family (Troglodytidae) and order (Passeriformes). Domain knowledge reveals that Marsh Wrens have a more distinctive eyebrow than House Wrens. 
In~\cref{fig:Cam}, the order level focuses on the head and legs, and the family level pays close attention to the head and feathers on the wing and tail. In addition to feathers, the species level concentrates on the head. 
Three hierarchical levels give similar attention to the head with a distinctive eyebrow, while they have unique regions of interest.

\begin{table*}
  \centering\footnotesize
  \caption{OA(\%)/$AU(\overline{PRC})$ results on \textbf{CUB-200-2011} by comparing to state-of-the-art methods.}
  \label{tab:sota_cub}
  \begin{tabular}{c|c|cc|cc|cc|cc|cc}
    \hline
    Relabeling & Hierarchy & \multicolumn{2}{|c|}{HMC-LMLP~\cite{cerri2016reduction}} & \multicolumn{2}{|c|}{HMCN~\cite{wehrmann2018hierarchical}} & \multicolumn{2}{|c|}{C-HMCNN~\cite{giunchiglia2020coherent}} & \multicolumn{2}{|c|}{Chang \etal~\cite{chang2021your}} & \multicolumn{2}{|c}{Ours} \\
    \hline
    \multirow{3}{*}{0\%} & Order & 98.45 & \multirow{3}{*}{0.945} & 97.29 & \multirow{3}{*}{0.934} & 98.48 & \multirow{3}{*}{0.960} & 97.76 & \multirow{3}{*}{0.968} & \textbf{98.67} & \multirow{3}{*}{\textbf{0.969}} \\
                         & Family & 94.24 &  & 93.15 &  & 94.63 &  & 94.17 &  & \textbf{95.51} & \\
                         & Specie & 79.60 &  & 79.75 &  & 81.58 &  & 85.56 &  & \textbf{86.60} & \\
    \hline
    \multirow{3}{*}{30\%} & Order & 98.17 & \multirow{3}{*}{0.920} & 96.82 & \multirow{3}{*}{0.905} & 97.98 & \multirow{3}{*}{0.938} & 97.81 & \multirow{3}{*}{\textbf{0.962}} & \textbf{98.31} & \multirow{3}{*}{0.958} \\
                         & Family & 93.58 &  & 91.99 &  & 93.89 &  & 94.10 &  & \textbf{94.79} & \\
                         & Specie & 71.30 &  & 71.68 &  & 74.91 &  & 82.53 &  & \textbf{83.91} & \\
    \hline
    \multirow{3}{*}{50\%} & Order & \textbf{98.36} & \multirow{3}{*}{0.895} & 96.70 & \multirow{3}{*}{0.874} & 98.34 & \multirow{3}{*}{0.909} & 97.43 & \multirow{3}{*}{\textbf{0.951}} & 97.89 & \multirow{3}{*}{0.944} \\
                         & Family & 93.84 &  & 90.85 &  & 94.10 &  & 93.47 &  & \textbf{94.29} & \\
                         & Specie & 64.34 &  & 64.29 &  & 67.52 &  & 79.30 &  & \textbf{80.52} & \\
    \hline
    \multirow{3}{*}{70\%} & Order & 98.27 & \multirow{3}{*}{0.831} & 97.22 & \multirow{3}{*}{0.834} & 98.02 & \multirow{3}{*}{0.844} & 96.65 & \multirow{3}{*}{0.924} & \textbf{98.43} & \multirow{3}{*}{\textbf{0.936}} \\
                         & Family & 93.84 &  & 91.25 &  & 93.91 &  & 91.74 &  & \textbf{93.94} & \\
                         & Specie & 47.98 &  & 52.90 &  & 50.05 &  & 70.03 &  & \textbf{73.96} & \\
    \hline
    \multirow{3}{*}{90\%} & Order & \textbf{98.38} & \multirow{3}{*}{0.716} & 97.31 & \multirow{3}{*}{0.725} & 98.27 & \multirow{3}{*}{0.772} & 97.12 & \multirow{3}{*}{\textbf{0.868}} & 97.97 & \multirow{3}{*}{0.865} \\
                         & Family & \textbf{94.44} &  & 86.85 &  & 94.37 &  & 91.91 &  & 93.32 & \\
                         & Specie & 22.89 &  & 30.69 &  & 26.16 &  & 49.36 &  & \textbf{53.02} & \\
    \hline
  \end{tabular}
\end{table*}

\begin{table*}
  \centering\footnotesize
  \caption{OA(\%)/$AU(\overline{PRC})$ results on \textbf{Aircraft} by comparing to state-of-the-art methods.}
  \label{tab:sota_air}
  \begin{tabular}{c|c|cc|cc|cc|cc|cc}
    \hline
    Relabeling & Hierarchy & \multicolumn{2}{|c|}{HMC-LMLP~\cite{cerri2016reduction}} & \multicolumn{2}{|c|}{HMCN~\cite{wehrmann2018hierarchical}} & \multicolumn{2}{|c|}{C-HMCNN~\cite{giunchiglia2020coherent}} & \multicolumn{2}{|c|}{Chang \etal~\cite{chang2021your}} & \multicolumn{2}{|c}{Ours} \\
    \hline
    \multirow{3}{*}{0\%} & Maker & 97.09 & \multirow{3}{*}{0.968} & 96.07 & \multirow{3}{*}{0.959} & \textbf{97.45} & \multirow{3}{*}{0.979} & 96.88 & \multirow{3}{*}{\textbf{0.981}} & \textbf{97.45} & \multirow{3}{*}{0.976} \\
                         & Family & 94.39 &  & 92.56 &  & 95.41 &  & 95.28 &  & \textbf{95.79} & \\
                         & Model & 90.25 &  & 87.19 &  & 91.69 &  & 91.92 &  & \textbf{92.58} & \\
    \hline
    \multirow{3}{*}{30\%} & Maker & 96.85 & \multirow{3}{*}{0.950} & 96.13 & \multirow{3}{*}{0.952} & 96.76 & \multirow{3}{*}{\textbf{0.971}} & 87.41 & \multirow{3}{*}{0.957} & \textbf{97.27} & \multirow{3}{*}{0.970} \\
                         & Family & 93.34 &  & 92.74 &  & 94.27 &  & 94.44 &  & \textbf{95.52} & \\
                         & Model & 85.42 &  & 85.42 &  & 88.39 &  & 89.33 &  & \textbf{91.62} & \\
    \hline
    \multirow{3}{*}{50\%} & Maker & 97.24 & \multirow{3}{*}{0.925} & 95.71 & \multirow{3}{*}{0.935} & 96.49 & \multirow{3}{*}{0.963} & 73.56 & \multirow{3}{*}{0.909} & \textbf{97.27} & \multirow{3}{*}{\textbf{0.965}} \\
                         & Family & 93.82 &  & 92.05 &  & 93.88 &  & 94.17 &  & \textbf{95.67} & \\
                         & Model & 83.59 &  & 81.52 &  & 85.18 &  & 86.66 &  & \textbf{89.66} & \\
    \hline
    \multirow{3}{*}{70\%} & Maker & 96.97 & \multirow{3}{*}{0.898} & 95.80 & \multirow{3}{*}{0.900} & 96.67 & \multirow{3}{*}{\textbf{0.953}} & 58.77 & \multirow{3}{*}{0.816} & \textbf{96.75} & \multirow{3}{*}{\textbf{0.953}} \\
                         & Family & 93.70 &  & 90.49 &  & 94.00 &  & 93.78 &  & \textbf{94.20} & \\
                         & Model & 81.61 &  & 78.37 &  & 80.11 &  & 82.96 &  & \textbf{84.53} & \\
    \hline
    \multirow{3}{*}{90\%} & Maker & \textbf{96.97} & \multirow{3}{*}{0.870} & 93.40 & \multirow{3}{*}{0.824} & 96.76 & \multirow{3}{*}{0.903} & 49.88 & \multirow{3}{*}{0.656} & 95.43 & \multirow{3}{*}{\textbf{0.904}} \\
                         & Family & 93.37 &  & 89.50 &  & \textbf{94.36} &  & 93.72 &  & 91.68 & \\
                         & Model & \textbf{74.41} &  & 70.06 &  & 71.02 &  & 64.99 &  & 71.06 & \\
    \hline
  \end{tabular}
\end{table*}

\subsection{Comparison with State-of-the-art Methods}\label{ssec:sota}

To fairly evaluate the proposed method, we compare it to state-of-the-art HMC methods: HMC-LMLP~\cite{cerri2016reduction}, HMCN~\cite{wehrmann2018hierarchical}, and C-HMCNN~\cite{giunchiglia2020coherent}, and the state-of-the-art FGVC approach that exploits the label hierarchy: Chang \etal~\cite{chang2021your}. In our hierarchical settings, we train all methods with different relabeling proportions. Chang \etal~\cite{chang2021your} sum the multi-class cross-entropy loss from each hierarchical level. When adapting their approach to hierarchical settings, we neglect the last-level loss if a sample has been relabeled to its parent class. We report OA of each hierarchical level and $AU(\overline{PRC})$ results on test sets of three FGVC datasets: CUB-200-2011, Aircraft, and Stanford Cars, displayed in \cref{tab:sota_cub}, \cref{tab:sota_air}, and \cref{tab:sota_car}, respectively.

From~\cref{tab:sota_cub}, \cref{tab:sota_air}, and \cref{tab:sota_car}, we can observe that the proposed method achieves the best OA results of each hierarchical level and the best $AU(\overline{PRC})$ results in most cases. In other cases, our results are also comparable to the best results. 
Chang \etal~\cite{chang2021your} use level-specific classification heads to disentangle coarse-level features with fine-grained ones, but they only consider mutually exclusion in each hierarchical level without examining subsumption relations between hierarchical levels in their loss function. C-HMCNN only constrains subsumption relations. 
HMC-LMLP and HMCN embed label hierarchy in their network architecture and train with the binary cross-entropy loss that implies all classes are independent. By contrast, in our framework, the tree hierarchy specifies the relation between any two labels with mutually exclusion or subsumption, and the corresponding probabilistic loss combined with the multi-class cross-entropy loss can transfer hierarchical knowledge during training. In addition, the proposed HRN disentangles hierarchical features by granularity-specific blocks, and these features interact via residual connections to fuse attributes following the hierarchy.

\begin{table*}
  \centering\footnotesize
  \caption{OA(\%)/$AU(\overline{PRC})$ results on \textbf{Stanford Cars} by comparing to state-of-the-art methods.}
  \label{tab:sota_car}
  \begin{tabular}{c|c|cc|cc|cc|cc|cc}
    \hline
    Relabeling & Hierarchy & \multicolumn{2}{|c|}{HMC-LMLP~\cite{cerri2016reduction}} & \multicolumn{2}{|c|}{HMCN~\cite{wehrmann2018hierarchical}} & \multicolumn{2}{|c|}{C-HMCNN~\cite{giunchiglia2020coherent}} & \multicolumn{2}{|c|}{Chang \etal~\cite{chang2021your}} & \multicolumn{2}{|c}{Ours} \\
    \hline
    \multirow{2}{*}{0\%} & Type & 96.98 & \multirow{3}{*}{0.953} & 95.21 & \multirow{3}{*}{0.938} & 96.75 & \multirow{3}{*}{0.971} & 96.40 & \multirow{3}{*}{0.977} & \textbf{97.41} & \multirow{3}{*}{\textbf{0.981}} \\
                         & Maker & 87.65 &  & 88.71 &  & 90.64 &  & 93.65 &  & \textbf{94.03} & \\
    \hline
    \multirow{2}{*}{30\%} & Type & \textbf{96.85} & \multirow{3}{*}{0.909} & 94.38 & \multirow{3}{*}{0.887} & 96.23 & \multirow{3}{*}{0.927} & 96.23 & \multirow{3}{*}{\textbf{0.970}} & 96.13 & \multirow{3}{*}{0.969} \\
                         & Maker & 79.16 &  & 81.59 &  & 81.92 &  & \textbf{91.61} &  & 90.55 & \\
    \hline
    \multirow{2}{*}{50\%} & Type & \textbf{96.92} & \multirow{3}{*}{0.842} & 93.46 & \multirow{3}{*}{0.832} & 95.95 & \multirow{3}{*}{0.850} & 95.60 & \multirow{3}{*}{0.960} & 95.88 & \multirow{3}{*}{\textbf{0.963}} \\
                         & Maker & 66.45 &  & 73.03 &  & 70.22 &  & 88.10 &  & \textbf{88.72} & \\
    \hline
    \multirow{2}{*}{70\%} & Type & \textbf{96.89} & \multirow{3}{*}{0.705} & 93.02 & \multirow{3}{*}{0.713} & 95.67 & \multirow{3}{*}{0.708} & 92.90 & \multirow{3}{*}{0.905} & 96.06 & \multirow{3}{*}{\textbf{0.947}} \\
                         & Maker & 41.52 &  & 52.66 &  & 43.17 &  & 76.13 &  & \textbf{83.72} & \\
    \hline
    \multirow{2}{*}{90\%} & Type & 96.38 & \multirow{3}{*}{0.572} & 93.42 & \multirow{3}{*}{0.560} & \textbf{96.49} & \multirow{3}{*}{0.577} & 92.25 & \multirow{3}{*}{0.761} & 94.32 & \multirow{3}{*}{\textbf{0.794}} \\
                         & Maker & 13.51 &  & 19.89 &  & 13.54 &  & 45.79 &  & \textbf{49.30} & \\
    \hline
  \end{tabular}
\end{table*}

\subsection{Analyze State-of-the-art Methods by Reducing Image Resolution}\label{ssec:reduce}

Except for domain knowledge, samples captured at low-resolution can hardly be identified with the last-level fine-grained categories, and thus they are more likely to be inferred as upper-level coarse classes. Considering the practical limitation of image quality, we reduce the image resolution of selected samples corresponding to different relabeling proportions. \cref{tab:sota_cub_ds} displays the experimental results, and our method consistently outperforms compared methods in most cases under two evaluation metrics. 

\begin{table*}
  \centering\footnotesize
  \caption{Compared OA(\%)/$AU(\overline{PRC})$ results on \textbf{CUB-200-2011} by reducing the image resolution after relabeling.}
  \label{tab:sota_cub_ds}
  \begin{tabular}{c|c|cc|cc|cc|cc|cc}
    \hline
    Relabeling & Hierarchy & \multicolumn{2}{|c|}{HMC-LMLP~\cite{cerri2016reduction}} & \multicolumn{2}{|c|}{HMCN~\cite{wehrmann2018hierarchical}} & \multicolumn{2}{|c|}{C-HMCNN~\cite{giunchiglia2020coherent}} & \multicolumn{2}{|c|}{Chang \etal~\cite{chang2021your}} & \multicolumn{2}{|c}{Ours} \\
    \hline
    \multirow{3}{*}{0\%} & Order & 98.45 & \multirow{3}{*}{0.945} & 97.29 & \multirow{3}{*}{0.934} & 98.48 & \multirow{3}{*}{0.960} & 97.76 & \multirow{3}{*}{0.968} & \textbf{98.67} & \multirow{3}{*}{\textbf{0.969}} \\
                         & Family & 94.24 &  & 93.15 &  & 94.63 &  & 94.17 &  & \textbf{95.51} & \\
                         & Specie & 79.60 &  & 79.75 &  & 81.58 &  & 85.56 &  & \textbf{86.60} & \\
    \hline
    \multirow{3}{*}{30\%} & Order & 97.86 & \multirow{3}{*}{0.926} & 96.32 & \multirow{3}{*}{0.887} & 97.81 & \multirow{3}{*}{0.944} & 97.62 & \multirow{3}{*}{\textbf{0.961}} & \textbf{98.50} & \multirow{3}{*}{0.959} \\
                         & Family & 93.18 &  & 88.06 &  & 93.48 &  & 93.59 &  & \textbf{94.75} & \\
                         & Specie & 74.32 &  & 70.78 &  & 76.04 &  & 82.33 &  & \textbf{84.13} & \\
    \hline
    \multirow{3}{*}{50\%} & Order & 97.45 & \multirow{3}{*}{0.907} & 95.32 & \multirow{3}{*}{0.853} & 97.62 & \multirow{3}{*}{0.925} & 97.12 & \multirow{3}{*}{0.947} & \textbf{98.20} & \multirow{3}{*}{\textbf{0.952}} \\
                         & Family & 92.25 &  & 85.93 &  & 92.51 &  & 91.79 &  & \textbf{93.82} & \\
                         & Specie & 68.10 &  & 62.70 &  & 70.37 &  & 78.30 &  & \textbf{81.18} & \\
    \hline
    \multirow{3}{*}{70\%} & Order & \textbf{97.62} & \multirow{3}{*}{0.862} & 94.43 & \multirow{3}{*}{0.789} & 97.20 & \multirow{3}{*}{0.881} & 96.32 & \multirow{3}{*}{0.909} & 97.58 & \multirow{3}{*}{\textbf{0.926}} \\
                         & Family & 91.72 &  & 82.64 &  & 91.18 &  & 88.84 &  & \textbf{92.42} & \\
                         & Specie & 53.11 &  & 45.75 &  & 55.14 &  & 68.06 &  & \textbf{73.98} & \\
    \hline
    \multirow{3}{*}{90\%} & Order & 96.67 & \multirow{3}{*}{0.695} & 93.56 & \multirow{3}{*}{0.694} & \textbf{96.79} & \multirow{3}{*}{0.801} & 96.06 & \multirow{3}{*}{\textbf{0.843}} & 96.15 & \multirow{3}{*}{0.837} \\
                         & Family & \textbf{89.96} &  & 78.60 &  & 89.44 &  & 87.52 &  & 88.29 & \\
                         & Specie & 20.78 &  & 22.52 &  & 28.32 &  & 46.58 &  & \textbf{50.10} & \\
    \hline
  \end{tabular}
\end{table*}

\subsection{Comparison with FGVC Methods Exploiting Hierarchical Knowledge}\label{ssec:fgvc}

Considering hierarchical knowledge, FGVC approaches refine the feature representation related to hierarchical levels in the feature space~\cite{zhang2016embedding,shi2018fine,chen2018fine,chang2021your}, \eg, measuring the distance between classes in the hierarchy~\cite{zhang2016embedding,shi2018fine}, learning finer-grained features with the prediction of higher level~\cite{chen2018fine}, or disentangling coarse-level features with fine-grained ones~\cite{chang2021your}. Nevertheless, they develop their loss functions based on the multi-class cross-entropy loss, which implies mutual exclusion at the same hierarchical level. On the other hand, encoding label relations like the parent-child correlation helps to utilize samples observed at different levels.
In contrast, the proposed method specifies label relations with the tree hierarchy and computes the combinatorial loss to effectively exploit samples labeled at different levels. 
 
We record the best results reported in their works in which each sample has complete hierarchical multi-granularity labels. 
In~\cref{tab:fgvc}, Chang \etal~\cite{chang2021your} achieve state-of-the-art performances in the traditional single-label FGVC problem. Our approach reaches comparable results by simply replacing ResNet-50 with ResNeXt101-32$\times$4d~\cite{xie2017aggregated}. Other techniques that enrich the feature representation in the context of FGVC can be applied to boost the performance, which is beyond our scope.

\begin{table}
  \centering\footnotesize
  \caption{OA on the species level by comparing to other FGVC approaches that exploit the label hierarchy.}
  \label{tab:fgvc}
  \begin{tabular}{cccc}
    \toprule
    Method & CUB-200-2011 & Aircraft & Stanford Cars \\
    \midrule
    Zhang \etal~\cite{zhang2016embedding} &  &  & 88.4 \\
    Shi \etal~\cite{shi2018fine} & 77.0 & 84.6 & 89.5 \\
    Chen \etal~\cite{chen2018fine} & 88.1 &  &  \\
    Chang \etal~\cite{chang2021your} & \textbf{89.9} & 93.6 & \textbf{95.1} \\
    Ours & 88.6 & \textbf{94.1} & \textbf{95.1} \\
    \bottomrule
  \end{tabular}
\end{table}



\section{Conclusion}\label{sec:con}
 
We study the HMC problem in which different objects can be discerned at various levels in the label hierarchy due to the differences in domain knowledge or image quality. To address this problem, we propose combinatorial loss and HRN. The combinatorial loss combines the probabilistic classification loss defined on the tree hierarchy that encodes semantic relations between any two hierarchical labels with the multi-class entropy loss imposed on the fine-grained leaf categories. The probabilistic classification loss can transfer hierarchical knowledge across levels, and the multi-class entropy loss increases the discriminative power on the leaf classes. In addition, HRN manages to perform hierarchical feature interaction via residual connections, \ie, features from parent levels acting as skip connections are added to features of children levels. Comprehensive experiments on three commonly used datasets demonstrated the effectiveness of the proposed method compared to state-of-the-art HMC and FGVC methods.


{\small
\bibliographystyle{ieee_fullname}
\bibliography{egbib}

\begin{thebibliography}{10}\itemsep=-1pt

\bibitem{barutcuoglu2006hierarchical}
Z. Barutcuoglu, R.~E. Schapire, and O.~G. Troyanskaya.
\newblock Hierarchical multi-label prediction of gene function.
\newblock {\em BMC Bioinform.}, 22(7):830--836, 2006.

\bibitem{cerri2014hierarchical}
R. Cerri, R.C. Barros, and A.~C. P. L.~F. de Carvalho.
\newblock Hierarchical multi-label classification using local neural networks.
\newblock {\em J. Comput. Syst. Sci.}, 80(1):39--56, 2014.

\bibitem{cerri2016reduction}
R. Cerri, R.~C. Barros, A.~C. P. L.~F. de Carvalho, and Y. Jin.
\newblock Reduction strategies for hierarchical multi-label classification in
  protein function prediction.
\newblock {\em BMC Bioinformat.}, 17(1):373, 2016.

\bibitem{chang2021your}
Dongliang Chang, Kaiyue Pang, Yixiao Zheng, Zhanyu Ma, Yi-Zhe Song, and Jun
  Guo.
\newblock Your" flamingo" is my" bird": Fine-grained, or not.
\newblock In {\em CVPR}, pages 11476--11485, 2021.

\bibitem{chen2020hyperbolic}
B. Chen, X. Huang, L. Xiao, Z. Cai, and L. Jing.
\newblock Hyperbolic interaction model for hierarchical multi-label
  classification.
\newblock In {\em AAAI}, volume~34, pages 7496--7503, 2020.

\bibitem{chen2018fine}
Tianshui Chen, Wenxi Wu, Yuefang Gao, Le Dong, Xiaonan Luo, and Liang Lin.
\newblock Fine-grained representation learning and recognition by exploiting
  hierarchical semantic embedding.
\newblock In {\em ACM MM}, pages 2023--2031, 2018.

\bibitem{deng2014large}
J. Deng, N. Ding, Y. Jia, A. Frome, K. Murphy, S. Bengio, Y. Li, H. Neven, and
  H. Adam.
\newblock Large-scale object classification using label relation graphs.
\newblock In {\em ECCV}, pages 48--64, 2014.

\bibitem{dimitrovski2011hierarchical}
I. Dimitrovski, D. Kocev, S. Loskovska, and S. D{z}eroski.
\newblock Hierarchical annotation of medical images.
\newblock {\em Pattern Recognit.}, 44(10-11):2436--2449, 2011.

\bibitem{dimitrovski2012hierarchical}
I. Dimitrovski, D. Kocev, S. Loskovska, and S. D{\v{z}}eroski.
\newblock Hierarchical classification of diatom images using ensembles of
  predictive clustering trees.
\newblock {\em Ecological Informat.}, 7(1):19--29, 2012.

\bibitem{giunchiglia2020coherent}
E. Giunchiglia and T. Lukasiewicz.
\newblock Coherent hierarchical multi-label classification networks.
\newblock In {\em NeurIPS}, volume~33, 2020.

\bibitem{he2016identity}
Kaiming He, Xiangyu Zhang, Shaoqing Ren, and Jian Sun.
\newblock Identity mappings in deep residual networks.
\newblock In {\em ECCV}, pages 630--645. Springer, 2016.

\bibitem{howard2019searching}
Andrew Howard, Mark Sandler, Grace Chu, Liang-Chieh Chen, Bo Chen, Mingxing
  Tan, Weijun Wang, Yukun Zhu, Ruoming Pang, Vijay Vasudevan, et~al.
\newblock Searching for mobilenetv3.
\newblock In {\em ICCV}, pages 1314--1324, 2019.

\bibitem{howard2017mobilenets}
Andrew~G Howard, Menglong Zhu, Bo Chen, Dmitry Kalenichenko, Weijun Wang,
  Tobias Weyand, Marco Andreetto, and Hartwig Adam.
\newblock Mobilenets: Efficient convolutional neural networks for mobile vision
  applications.
\newblock {\em arXiv preprint arXiv:1704.04861}, 2017.

\bibitem{huang2017densely}
Gao Huang, Zhuang Liu, Laurens Van Der~Maaten, and Kilian~Q Weinberger.
\newblock Densely connected convolutional networks.
\newblock In {\em CVPR}, pages 4700--4708, 2017.

\bibitem{silla2011survey}
C.~N.~Silla Jr and A.~A. Freitas.
\newblock A survey of hierarchical classification across different application
  domains.
\newblock {\em Data Min. Knowl. Discov.}, 22(1-2):31--72, 2011.

\bibitem{krause20133d}
Jonathan Krause, Michael Stark, Jia Deng, and Li Fei-Fei.
\newblock 3d object representations for fine-grained categorization.
\newblock In {\em CVPR workshops}, pages 554--561, 2013.

\bibitem{lewis2004rcv1}
D.~D. Lewis, Y. Yang, T.~G. Rose, and F. Li.
\newblock Rcv1: A new benchmark collection for text categorization research.
\newblock {\em J. Mach. Learn. Res.}, 5(Apr):361--397, 2004.

\bibitem{loshchilov2016sgdr}
Ilya Loshchilov and Frank Hutter.
\newblock Sgdr: Stochastic gradient descent with warm restarts.
\newblock {\em arXiv preprint arXiv:1608.03983}, 2016.

\bibitem{ma2018shufflenet}
Ningning Ma, Xiangyu Zhang, Hai-Tao Zheng, and Jian Sun.
\newblock Shufflenet v2: Practical guidelines for efficient cnn architecture
  design.
\newblock In {\em ECCV}, pages 116--131, 2018.

\bibitem{maji2013fine}
Subhransu Maji, Esa Rahtu, Juho Kannala, Matthew Blaschko, and Andrea Vedaldi.
\newblock Fine-grained visual classification of aircraft.
\newblock {\em arXiv preprint arXiv:1306.5151}, 2013.

\bibitem{mao2019hierarchical}
Y. Mao, J. Tian, J. Han, and X. Ren.
\newblock Hierarchical text classification with reinforced label assignment.
\newblock In {\em EMNLP-IJCNLP}, pages 445--455, 2019.

\bibitem{mayne2009hierarchically}
A. Mayne and R. Perry.
\newblock Hierarchically classifying documents with multiple labels.
\newblock In {\em Proc. IEEE Symp. Comput. Intell. Data Mining}, pages
  133--139. IEEE, 2009.

\bibitem{meng2019weakly}
Y. Meng, J. Shen, C. Zhang, and J. Han.
\newblock Weakly-supervised hierarchical text classification.
\newblock In {\em AAAI}, volume~33, pages 6826--6833, 2019.

\bibitem{peng2018large}
H. Peng, J. Li, Y. He, Y. Liu, M. Bao, L. Wang, Y. Song, and Q. Yang.
\newblock Large-scale hierarchical text classification with recursively
  regularized deep graph-cnn.
\newblock In {\em WWW}, pages 1063--1072, 2018.

\bibitem{rousu2006kernel}
J. Rousu, C. Saunders, S. Szedmak, and J. Shawe-Taylor.
\newblock Kernel-based learning of hierarchical multilabel classification
  models.
\newblock {\em J. Mach. Learn. Res.}, 7:1601--1626, 2006.

\bibitem{schietgat2010predicting}
L. Schietgat, C. Vens, J. Struyf, H. Blockeel, D. Kocev, and S. D{\v{z}}eroski.
\newblock Predicting gene function using hierarchical multi-label decision tree
  ensembles.
\newblock {\em BMC Bioinformat.}, 11(1):2, 2010.

\bibitem{selvaraju2017grad}
Ramprasaath~R Selvaraju, Michael Cogswell, Abhishek Das, Ramakrishna Vedantam,
  Devi Parikh, and Dhruv Batra.
\newblock Grad-cam: Visual explanations from deep networks via gradient-based
  localization.
\newblock In {\em ICCV}, pages 618--626, 2017.

\bibitem{shi2018fine}
Weiwei Shi, Yihong Gong, Xiaoyu Tao, De Cheng, and Nanning Zheng.
\newblock Fine-grained image classification using modified dcnns trained by
  cascaded softmax and generalized large-margin losses.
\newblock {\em NeurIPS}, 30(3):683--694, 2018.

\bibitem{huang2019hierarchical}
W.~Huang \textit{et al.}
\newblock Hierarchical multi-label text classification: An attention-based
  recurrent network approach.
\newblock In {\em Proc. 28th ACM Int. Conf. Inf. Knowl. Manag.}, pages
  1051--1060, 2019.

\bibitem{vaswani2017attention}
Ashish Vaswani, Noam Shazeer, Niki Parmar, Jakob Uszkoreit, Llion Jones,
  Aidan~N Gomez, {\L}ukasz Kaiser, and Illia Polosukhin.
\newblock Attention is all you need.
\newblock In {\em NeurIPS}, pages 5998--6008, 2017.

\bibitem{vens2008decision}
C. Vens, J. Struyf, L. Schietgat, S. D{\v{z}}eroski, and H. Blockeel.
\newblock Decision trees for hierarchical multi-label classification.
\newblock {\em Mach. Learn.}, 73(2):185--214, 2008.

\bibitem{wah2011caltech}
Catherine Wah, Steve Branson, Peter Welinder, Pietro Perona, and Serge
  Belongie.
\newblock The caltech-ucsd birds-200-2011 dataset.
\newblock 2011.

\bibitem{wehrmann2018hierarchical}
J. Wehrmann, R. Cerri, and R.C. Barros.
\newblock Hierarchical multi-label classification networks.
\newblock In {\em ICML}, pages 5075--5084, 2018.

\bibitem{xie2017aggregated}
Saining Xie, Ross Girshick, Piotr Doll{\'a}r, Zhuowen Tu, and Kaiming He.
\newblock Aggregated residual transformations for deep neural networks.
\newblock In {\em CVPR}, pages 1492--1500, 2017.

\bibitem{zhang2016embedding}
Xiaofan Zhang, Feng Zhou, Yuanqing Lin, and Shaoting Zhang.
\newblock Embedding label structures for fine-grained feature representation.
\newblock In {\em CVPR}, pages 1114--1123, 2016.

\bibitem{zhou2016fine}
Feng Zhou and Yuanqing Lin.
\newblock Fine-grained image classification by exploring bipartite-graph
  labels.
\newblock In {\em CVPR}, pages 1124--1133, 2016.

\end{thebibliography}
}

\end{document}